\pdfoutput=1
\documentclass[11pt]{article}

\usepackage{ACL2023}

\usepackage{times}
\usepackage{latexsym}

\usepackage[T1]{fontenc}
\usepackage[utf8]{inputenc}

\usepackage{microtype}
\usepackage{times}
\usepackage{latexsym}
\usepackage[T1]{fontenc}
\usepackage[utf8]{inputenc}
\usepackage{tabularx,booktabs}

\usepackage{microtype}

\usepackage{graphicx}
\usepackage{eurosym}
\usepackage{tablefootnote}
\graphicspath{ {images/} }
\usepackage{hyperref}
\usepackage{textcomp}
\usepackage{hhline}
\usepackage[T1]{fontenc}
\usepackage{listings}
\usepackage{stfloats}
\usepackage{csquotes}
\usepackage{arydshln}
\usepackage{booktabs}
\usepackage{array}
\usepackage[super]{nth}
\usepackage[inline,shortlabels]{enumitem}
\usepackage{multirow}
\usepackage{makecell}
\usepackage{siunitx}

\usepackage{multicol}
\usepackage{svg}
\usepackage{float}
\usepackage{amssymb}
\usepackage{amsmath,amsfonts,amssymb}
\usepackage[caption=false]{subfig}
\usepackage{graphicx}

\usepackage{inconsolata}

\newcommand{\LLM}{PaLM 2}

\title{Language and Task Arithmetic with Parameter-Efficient Layers \\ 
for Zero-Shot Summarization}

\author{Alexandra Chronopoulou$^{1}$\thanks{\hspace{1mm} Work done during an internship at Google DeepMind. Correspondence to \texttt{alexandrachron@google.com}} \quad Jonas Pfeiffer$^{2}$ \quad Joshua Maynez$^{2}$  \\ 
 \quad \textbf{Xinyi Wang}$^{2}$ \quad \textbf{Sebastian Ruder}$^{3} $\thanks{\hspace{1mm} Work done while working at Google} \quad \textbf{Priyanka Agrawal}$^{2}$ \\
 $^{1}$Google \quad  $^{2}$Google DeepMind \quad $^{3}$Cohere \\ 
 	{\tt } \\ 
  }
\begin{document}
\maketitle

\begin{abstract}

Parameter-efficient fine-tuning~(PEFT) using labeled task data can significantly improve the performance of large language models~(LLMs) on the downstream task. However, there are 7000 languages in the world and many of these languages lack labeled data for real-world language generation tasks. 
In this paper, we propose to improve zero-shot cross-lingual 
transfer by composing expert modules trained separately on language or task data.
Our method composes \textit{language} and \textit{task} PEFT adapters via element-wise arithmetic operations to leverage unlabeled data and English labeled data.
We extend our approach to cases where labeled data from more languages is available and propose to arithmetically compose PEFT adapters trained on languages related to the target.
Empirical results on summarization demonstrate that our method is a strategy that obtains consistent gains using minimal training of PEFT parameters.

\end{abstract}

\section{Introduction}
Large language models~(LLM) have achieved impressive performance on various real world applications in many different human languages~\cite{xue-etal-2021-mt5, brown, chowdhery2022palm,palm2, jiang2024mixtralexperts}. 
Summarization~\cite{INR-015} is a particularly interesting and useful task because it allows users to quickly aggregate and access relevant information from large amounts of textual data. 
Developing a competitive text summarization system for a language typically involves fine-tuning a pretrained model on labeled summarization data in the given language. Standard supervised fine-tuning of LLMs can be very expensive due to the large model size. Parameter-efficient tuning~(PEFT) is an effective alternative that achieves competitive performance while incurring much less computational and memory cost~\citep{hu2022lora,lester-etal-2021-power,zhang2023llamaadapterefficientfinetuninglanguage}.

Despite the effectiveness of PEFT~\citep{touvron2023llama}, it also has several limitations if we want to develop competitive multilingual summarization systems. First, current PEFT methods generally require access to labeled task data in a given language. While there are several existing datasets in English to train competitive summarization systems~\cite{NIPS2015_afdec700, grusky-etal-2018-newsroom, narayan-etal-2018-dont}, many languages in the world with millions of speakers do not have such resources~\cite{giannakopoulos-etal-2015-multiling,scialom-etal-2020-mlsum, Cao_Wan_Yao_Yu_2020}. Second, standard PEFT methods optimize a separate set of parameters for each language, resulting in thousands of fine-tuned checkpoints, which need to be stored and deployed individually \cite{NEURIPS2021_e77910eb}. Finally, as the standard PEFT methods are fine-tuned in isolation, they cannot leverage information from related tasks. 
%which could be impractical if we want to scale to thousands of human languages in the world.   

In this paper, we want to improve zero-shot multilingual summarization with PEFT to better support languages that might lack labeled summarization data. To this end, we propose a simple yet effective method that composes language and task information stored in different trained PEFT parameters through element-wise operation. We leverage unlabeled data to train language parameters with PEFT, and perform element-wise arithmetic operations with pretrained \textit{task} and \textit{language} parameters to construct new parameters for a language without labeled summarization data. While several prior works have studied methods that compose PEFT methods for zero-shot cross-lingual transfer~\citep{pfeiffer-etal-2020-mad,vu-etal-2022-overcoming}, these methods generally incur an additional inference cost. Our method provides a simpler and more flexible framework to leverage many related languages at a fixed inference cost.     

Our method is inspired by the lottery ticket hypothesis~\cite{frankle2018the}, which posits that distinct models fine-tuned on the same dataset follow linear trajectories while maintaining a consistent loss~\cite{frankle2020, yunis2022on}. This hypothesis implies that element-wise operations on different fine-tuned models can also remove biases of the pretrained model \cite{ilharco2023editing}, allowing the accumulation of information from auxiliary tasks \cite{fisheravg}, or improve adaptation to unforeseen textual domains \cite{li2022branchtrainmerge, chronopoulou-etal-2023-adaptersoup}. Our work is the first to extend this observation to improve cross-lingual transfer by combining pretrained language and task parameters.

Our contributions are the following:

\begin{figure*}[t]
	\centering
	\includegraphics[width=0.95\textwidth, page=1]{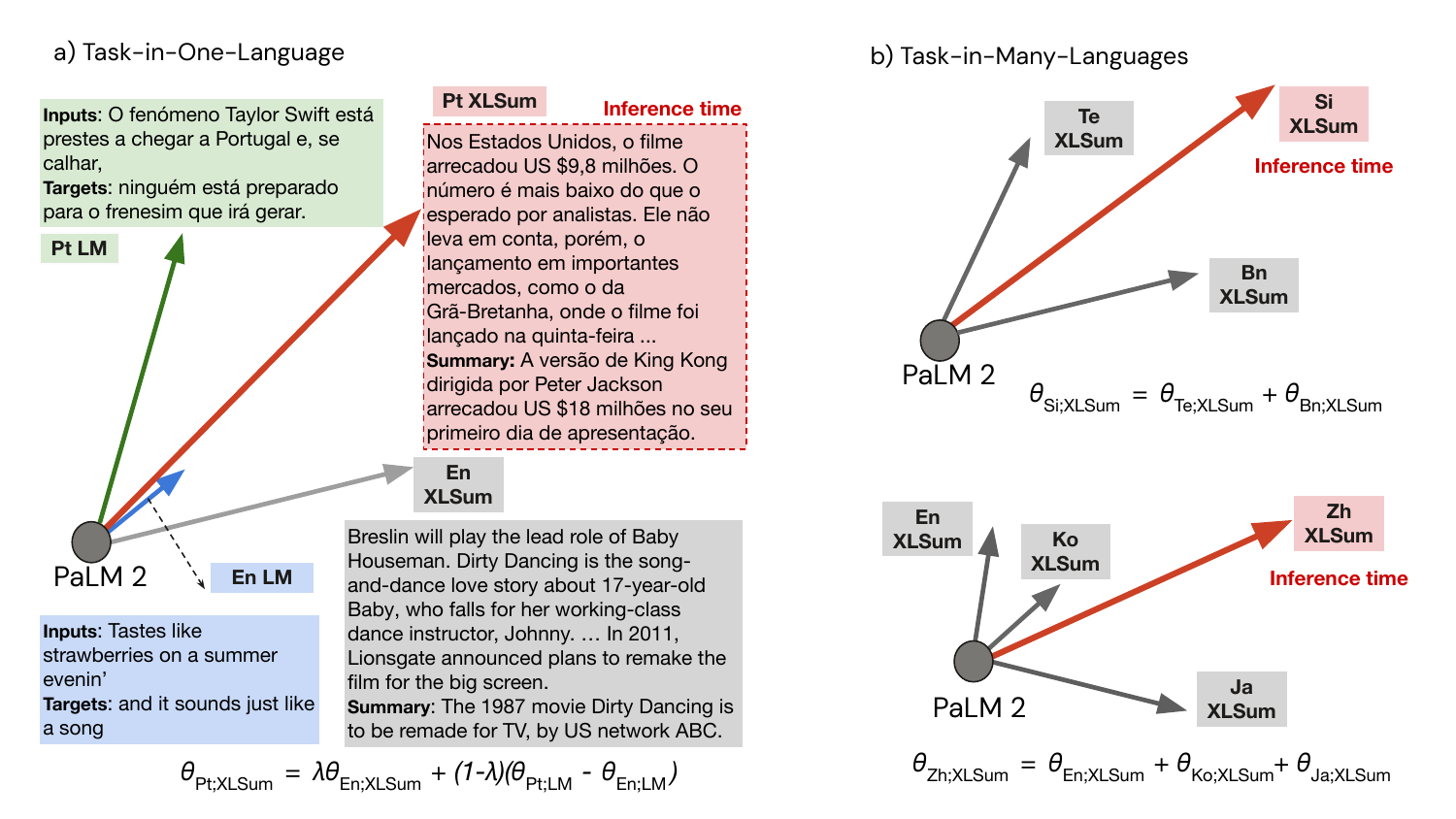}
        	\caption{\textbf{Illustration of our language and task arithmetic approach for zero-shot cross-lingual transfer using LoRA parameters learned on top of \LLM{}}. (a) We train a task adapter using the summarization objective in En and language adapters using Prefix-LM in En and Pt. At inference time, a summary is generated in Pt, shown with a dotted frame (Subsection \ref{subsection:arithmeticlangtask}). (b) We add the weights of task adapters trained for summarization in languages similar to the target. We use the resulting vector for zero-shot summarization in the target language (Subsection \ref{subsection:arithmetictask}).}
	\label{fig:joint}
\end{figure*}

\begin{enumerate}
\setlength{\itemsep}{0pt}
\item Assuming we only have task data in English, we combine PEFT parameters trained on English task data and unlabeled data in other languages through element-wise composition. This setup, termed \textit{Task-in-One-Language}, improves the model's summarization performance across all unseen target languages, as demonstrated on the XLSum benchmark~\citep{hasan-etal-2021-xl}.

\item Extending our first approach, we consider scenarios with task data from multiple languages (\textit{Task-in-Many-Languages}). When labeled task data for summarization are available in various languages, we combine representations from languages most related to the target, consistently improving performance over the baselines using the XLSum benchmark.

\item We apply our language and  task arithmetic to a different PEFT method, the Kronecker adapter \cite{edalati2022krona} and evaluate  its performance on XLSum and TyDi-QA \cite{clark-etal-2020-tydi}. We find that our approach is also effective with these other methods and tasks.
\end{enumerate}

\section{Language and Task Arithmetic}

Prior work has applied element-wise operations to the weights of fine-tuned models \cite{fisheravg, pmlr-v162-wortsman22a,ilharco2023editing, ainsworth2023git,yadav2023tiesmerging}, or PEFT modules \cite{chronopoulou-etal-2023-adaptersoup, zhang2023composing}. 
These studies demonstrate that interpolating the weights of fine-tuned models (or specific layers) effectively creates multi-task and multi-domain models. 
We hypothesize that element-wise operations can also be used to combine knowledge acquired in different languages. Our work is the first to propose the arithmetic composition of language and task PEFT modules for cross-lingual natural language generation. Figure 1 illustrates an overview of our approach. \ref{fig:joint}.

Our goal is to enable Large Language Models (LLMs) to support summarization in an unseen target language ($T$) for which we lack labeled data.  We assume access to labeled task data in other languages, as well as unlabeled monolingual data in both the source language ($S$) and the target language ($T$). In particular, we can use either labeled or unlabeled data to train small PEFT modules that capture the attributes of a given task or language. 

\noindent \textbf{Task Adapter:} We fine-tune an LLM using LoRA adapters on labeled data from XLSum  \cite{hasan-etal-2021-xl} in the source language $S$.  We refer to the fine-tuned model as Task Adapter. 
 
\noindent \textbf{Language Adapter:} We fine-tune LoRA parameters with LLMs on monolingual data in the source or target language ($S$ or $T$). We refer to the fine-tuned model as language adapter. We use the prefix-LM pretraining objective from T5 \cite{c4} with mC4 data to train language adapters. 

We propose to compose the \textit{language} and \textit{task} 
 vectors to better support summarization into the target language $T$. Next, we introduce our method under two different data settings.

\subsection{Task-in-One-Language}
\label{subsection:arithmeticlangtask}
First, we consider the zero-shot setting where the source language $S$ is English. We have labeled data in $S$, and some amount of unlabeled data both in the source language $S$ and the target language $T$.
\paragraph{Composing via Language and Task Addition:} 
\label{subsubsection:add}

We want to encourage the model to generate in the target language $T$ and learn the task from the data available in the source language $S$. 

Let $\theta_{\text{LM;T}}$ be the LoRA parameters trained on the monolingual data in the target language $T$, and $\theta_{\text{task;S}}$ be the LoRA parameters trained on the labeled task data in the source language $S$, we propose to calculate the zero-shot task module for the target language $T$ as:
\begin{align}
\theta_{\text{task;T}} 
= \lambda \theta_{\text{task;S}} +  
(1-\lambda)(\theta_{\text{LM;T}})  
\end{align}
The scaling term $\lambda$ is determined using held-out validation data. We refer to this approach as \textit{Language and Task; Add}.

\paragraph{Composing via Language and Task Addition and Subtraction:} 
\label{subsubsection:addsubtract}
We want to steer the model's ability to generate in the target language $T$, but avoid generating in the source language $S$. 
Previous work showed that subtraction can be a method of ``unlearning'' information \cite{ilharco2023editing,zhang2023composing}. 
% , we extrapolate this property to a cross-lingual setup. 
We propose \textit{subtracting} the source language adapter from the target language adapter. The intuition is that by negating the source language adapter, we control the generation, making the model ``forget'' the source language. 

Our goal in this zero-shot transfer setup is to obtain a model that has a \textbf{strong summarization ability} (learned from the task adapter) \textbf{in the correct target language} (learned from the target language adapter) \textbf{while \textit{not} generating in the source language} (unlearned from the source language adapter). 

Formally, let $\theta_{\text{LM;S}}$ be the LoRA parameters trained on the monolingual data in the source language $S$. We propose to calculate the zero-shot task module for the target language $T$ as:
\begin{align}
\label{eqn:addsub_single_lang}
\theta_{\text{task;T}} 
= \lambda \theta_{\text{task;S}} +  
(1-\lambda)(\theta_{\text{LM;T}} - \theta_{\text{LM;S}}) 
\end{align}
 where $\lambda$ is a hyperparameter tuned in the same way as in the previous setting. We refer to it as \textit{Language and Task; Add and Subtract}.
 
\subsection{Task-in-Many-Languages}
\label{subsection:arithmetictask}

Subsection \ref{subsection:arithmeticlangtask} presents language and task arithmetic when we want to do zero-shot transfer from a single source language $S$. However, in practice, we sometimes have data in many different source languages. In this subsection, we extend our language and task arithmetic framework to the setting where we utilize data in many different languages. 

\paragraph{Composing via Task-only Addition:}
First, we want to utilize labeled task data in various source languages.
Formally, given labeled task data for $N$ languages $(S_1, ..., S_N)$, we want to use the LLM to support an unseen target language $T$, for which we have no task data. 
To this end, given LoRA parameters $(\theta_{\text{task;S}_1}, ..., \theta_{\text{task;S}_N})$ trained on labeled task data in $(S_1, ..., S_N)$, we propose to perform zero-shot generation on the target language $T$ using the average of PEFT modules of its related languages:
\begin{align}
\label{eqn:add_many_lang}
    \theta_{\text{task;T}} = \frac{1}{L} \sum_{i=1}^L \theta_{\text{task;S}_i} 
\end{align}
 where $L$ <= $N$. 
If $L = N$, we essentially add the weights of all available task adapters (we name this method \textit{Task-only; Add all}). To select a subset of $L$ languages that are most related to the target language $T$, we use the URIEL language vectors~\cite{littell-etal-2017-uriel}. We retrieve the pre-computed syntactic and geographic distances between $T$ and each of the $N$ languages of the training set using an implementation of the toolkit lang2vec.\footnote{\url{https://github.com/antonisa/lang2vec}} We refer to this approach as \textit{Task-only; Add related}.

\paragraph{Composing via Language and Task Addition and Subtraction:}

Similarly, if we have both labeled and unlabeled data in several source languages, we can modify \autoref{eqn:addsub_single_lang} to leverage both types of data in many different languages:
\begin{align}
\theta_{\text{task;T}} 
= \lambda {\theta}^\prime_{\text{task;S}} +  
(1-\lambda)(\theta_{\text{LM;T}} - {\theta}^\prime_{\text{LM;S}}) 
\end{align} 

Where $\theta^\prime_{task;S} = \frac{1}{L} \sum_{i=1}^L \theta_{\text{task;S}_i} $ (as computed in \autoref{eqn:add_many_lang}), 
% \priyanka{isn't this same as eq (3)? We should refer it here} 
i.e., it is the average of the related (to the target $T$) task adapters, and $\theta^\prime_{LM;S} = \frac{1}{L} \sum_{i=1}^L \theta_{\text{LM;S}_i} $, i.e., it is the average of the related language adapters according to URIEL. This approach is denoted as \textit{Language and Task; Add and Subtract related}.

\begin{table*}[h]
\resizebox{0.95\textwidth}{!}{
\begin{tabular}{lccccccccccccc}
\toprule
  \textbf{Method}  & Mr  &  Gu & Zh & Ne & Pt & Si & So & Vi & Yo & Uk & Fa &  \multicolumn{1}{c}{Avg} \\
  \midrule 
 \multicolumn{5}{l}{\textbf{Task-in-One-Language}} & & &&&& & \\ 
Baseline & 20.5 & 30.3 & 23.9  & 29.4 & 22.3 & 34.5 & 21.3 & 24.5  & 17.3  & 17.4  & 25.1  & 24.2  \\
\midrule
% \multicolumn{5}{l}{{Language and Task Arithmetic}} & & &&&& & \\ 
Language and Task (Add) & 20.6 & 30.3 & 24.1 & 29.4 & 22.3 & 34.7 & 21.5 & 24.5  & 17.7 & 18.1 & 25.2 & 24.4  \\
Language and Task (Add and Subtract) & \textbf{20.7} & \textbf{30.6} & \textbf{24.6} & \textbf{29.6}  & \textbf{22.5}  & \textbf{35.4 }& \textbf{21.8}  & 24.6 & \textbf{18.5}  & \textbf{20.9 }&  \textbf{25.8} & \textbf{25.0}   \\

 % \multicolumn{4}{l}{\textbf{Zero-shot approaches (using only Ja labeled data)}} && & & &&&& & \\ 
% \hspace{4mm} - Ja LoRA & 18.9 & 28.3 & 23.9 & 27.6 & 21.8 & 34.2 & 21.5 & 23.5 & 17.1 & 20.4 & 23.7 & 24.1  \\
% \hspace{4mm} - Lora lang-task arithmetic (ours) & \textbf{19.3} & 28.3 & \textbf{24.6} & \textbf{28.2} & \textbf{22.0} & \textbf{34.5} & \textbf{21.7} & \textbf{23.8} & \textbf{17.9} & \textbf{20.7} & \textbf{24.0} & \textbf{24.5}  \\ \midrule
%  \multicolumn{4}{l}{\textbf{\textbf{Oracle approaches (using target language labeled data)}}} & & & &&&& & \\ 
% \hspace{2mm} - task adapter & 24.2 & 32.5 & 27.5 & 31.5 & 24.4 & 36.3 & 23.5 & 26.2 & 27.6 & 22.9 & 27.6  & 27.7 \\ 
% \hspace{2mm} - Full fine-tuning & 24.7 & 33.1 & 27.6 & 32.5 & 24.7 & 37.5 & 24.3 & 26.7 & 29.2 & 23.3 & 27.7 & 28.3 \\
\bottomrule 
\end{tabular} 

} \caption{\textbf{Language and task arithmetic improves zero-shot cross-lingual transfer on XLSum when we only have task data in En}. We show ROUGE-2 spm scores on $\text{XLSum}_{unseen}$. We train the task adapter using En $\text{XLSum}$ data and the language adapter using Prefix-LM on mC4 data.} \label{table:arithmeticresults}
\end{table*}

\begin{table*}[h]
\resizebox{0.95\textwidth}{!}{
\begin{tabular}{lrrrrrrrrrrrr}
\toprule
 \textbf{Method}  & Mr & Gu & Zh & Ne & Pt & Si & So & Vi & Yo & Uk & Fa &  \multicolumn{1}{c}{Avg} \\
\midrule 
\multicolumn{5}{l}{\textbf{Task-in-Many-Languages}} & & &&&& & \\ 
Baseline (best) & 21.2 & 31.2 & 25.6 & 28.4 & 22.5 & 35.8 & 22.1 & 25.6 & 21.4 & 21.6 & 25.3 & 25.5 \\
Baseline (multilingual) & \textbf{21.4} & 31.2 & \textbf{26.4} & 28.8 & 22.8 & 35.4 & 22.4 & \textbf{25.7} & 20.2 & 21.5 & 25.5 & 25.6 \\
\midrule
% \multicolumn{5}{l}{{Language and Task Arithmetic}} & & &&&& & \\
Task-only (Add all)& \textbf{21.4} & 31.3 & 25.6 & 28.6 & 22.8 & 35.4 & 22.0 & 25.5 & 20.4 & 21.3 & 25.5 & 25.4 \\
Task-only (Add related) & 21.1 & \textbf{31.5} & 25.4  & \textbf{30.2}  & 
\textbf{23.1} & \textbf{36.3} & \textbf{22.9} & 25.1 & \textbf{22.9} & \textbf{21.8} & \textbf{25.7} & \textbf{26.0} \\% \midrule 
% \multicolumn{8}{l}{{\textbf{Labeled data in non-target languages \& unlabeled data in all languages}}} & & & & \\ 
Language and Task (Add and Subtract related) & 21.2 & \textbf{31.5} & 25.4  & \textbf{30.4}  & 
\textbf{23.0} & \textbf{36.4} & \textbf{22.8} & 25.0 & \textbf{22.9} & \textbf{21.7} & \textbf{25.7} & \textbf{26.0} \\ 
% \midrule 
% \parbox[t]{2mm}{\multirow{4}{*}{\rotatebox[origin=c]{90}{Kronecker}}}  & Monolingual best & 21.3 & 31.4 & 25.6 & 30.0 & 22.6 & 36.0 & 22.9 & 25.4 & 21.8 & 22.0 & 25.7 & 25.9 \\
% & Multilingual & 21.2 & 31.5 &  26.1 & 30.8  & 23.2 & 36.7 & 23.1 & 25.5 & 21.5 & 22.0 & 25.9 & 26.1 \\
% & Weight avg & 20.9 & 31.3  &  25.6 & 30.5 & 22.8  &35.9  & 22.7 & 25.2 & 20.8 & 21.9 & 25.7 &  25.7 \\ 
% & Weight avg (related)&  21.1 &  32.2 & 26.2 & 31.4 & 24.0 & 36.6 & 22.9 & 25.7 & 21.9 & 22.3 & 26.6 & 26.4 \\ \midrule 
% \multicolumn{7}{l}{\textbf{\textbf{Oracle approaches (using target language labeled data)}}} &  &&&& & \\ 
% & LoRA  & 24.2 & 32.5 &27.5 &  31.5& 24.4 & 36.3 & 23.5 & 26.2 & 28.3 & 22.9 & 27.4 & 27.7 \\ 
% & Kronecker  & 24.3 & 32.8  & 27.8  & 31.9 & 24.2 & 35.0 & 24.0 & 26.4 & 28.1 & 23.2 & 27.7 & 27.8 \\ 

\bottomrule   
\end{tabular}

} \caption{\textbf{Addition of task adapters improves zero-shot cross-lingual transfer on XLSum when we have task data in multiple languages.} We show ROUGE-2 spm zero-shot scores on $\text{XLSum}_{unseen}$. 
}\label{table:taskinmanylangs}
\end{table*}

\section{Experimental Setup}
\label{sec:experiments}
\subsection{Tasks and Datasets} 

\noindent \textbf{{Summarization}}:
We use XLSum \cite{hasan-etal-2021-xl}, a news summarization dataset of BBC articles, where each article has a one-sentence summary. While prior work studies the zero-shot learning setting where only English labeled data is available~\cite{vu-etal-2022-overcoming}, we utilize the available multilingual training data for a more realistic setting. Specifically, we use a subset of  XLSum as our training set, and specifically the articles and summaries of the languages: Arabic ({ar}), Bengali (bn), English (en), Japanese (ja), Korean (ko), Indonesian (id), Swahili ({sw}), Russian ({ru}), Telugu ({te}), Thai ({th}), and Turkish ({tr}). We refer to this set as $\text{XLSum}_{seen}$. Training dataset stats are shown in Table \ref{table:xlsumseen} of the Appendix.

For zero-shot evaluation, we select 11 languages from XLSum as unseen languages: Marathi ({mr}), Gujarati ({gu}), Chinese simplified ({zh}), Nepali ({ne}), Portuguese ({pt}), Sinhala ({si}), Somali ({so}), Vietnamese ({vi}), Yoruba ({yo}), Ukrainian ({uk}), and Persian ({fa}). We do not use training data from any of these languages. We  refer to this set of 11 languages as $\text{XLSum}_{unseen}$. 

\noindent \textbf{Unlabeled data}:
We use unlabeled data from mC4 \cite{xue-etal-2021-mt5} with the prefix language modeling objective from T5 \cite{c4}. This corpus has been created using a Common Crawl-based dataset covering 101 languages. All languages considered in our experiments are covered by mC4. For the language adapters, we fine-tune the LLM using LoRA on prefix-LM for $5k$ steps in  each language.

\subsection{Training and Implementation Details} We
 use \LLM{}-S ~\cite{palm2}, a state-of-the-art, highly multilingual language model, as the base LLM for all our experiments. 
 
We add LoRA parameters of rank 4 to the Key, Query, Value, Projection  attention matrices. We do not tune this hyperparameter. This results in adding parameters that account for just 0.2\% of the parameters of \LLM{} (we do not update the weights of the pretrained model).  We fine-tune \LLM{} on prefix-LM, XLSum using LoRA with learning rate $2e-4$. 

For XLSum, we report ROUGE-2 \cite{lin-2004-rouge} as the evaluation metric for En, and SentencePiece-ROUGE-2 for all other languages. This is an extension of ROUGE that handles non-Latin character using a SentencePiece tokenizer; in this work, we use the mT5 tokenizer \cite{xue-etal-2021-mt5}.

\subsection{Baselines} 

\noindent {\textsc{\textbf{Task-in-One-Language:}}} The baseline is computed by fine-tuning \LLM{} on En XLSum data using LoRA parameters. During fine-tuning, only the LoRA parameters are being updated, while the underlying LLM remains frozen. 

\noindent{\textsc{\textbf{Task-in-Many-Languages:}}} 
The baseline is computed by fine-tuning \LLM{} on XLSum data of each of the language in $\text{XLSum}_{seen}$ independently using LoRA parameters. Then, the best-performing model (per target language) is selected. We denote this as \textit{baseline (best)}.

We also compute a \textit{multilingual baseline}: we simply concatenate the datasets of the different languages of $\text{XLSum}_{seen}$ and we train the LLM with LoRA on the entire dataset.\footnote{We also ran the full fine-tuning baselines and we observed that the gap to the PEFT baselines is small, results are shown in the Appendix.}

%Additionally, we fine-tune \LLM{} both in a standard  and a parameter-efficient way in each of the target languages ($\text{XLSum}_{unseen}$). These oracle approaches help us quantify the gap between zero-shot evaluation and supervised fine-tuning of \LLM{} on the task of summarization. 
% \priyanka{Baselines should be elaborated here}\alex{how does it look now?}

\section{Results and Discussion}

\subsection{Task-in-One-Language}

\noindent\textbf{Language and task arithmetic (Add and Subtract) improves zero-shot cross-lingual transfer:} 
We present the main results of our language and task arithmetic approach in cross-lingual summarization in Table \ref{table:arithmeticresults}. In the second row, we show the results by composing the language and task LoRA parameters via addition (\textit{language and task; add}). This approach provides only slight improvements over the task adapter baseline in terms of ROUGE-2. Our language and task arithmetic approach with addition and subtraction  (third row) consistently outperforms the baseline as well as the simple addition of source task and target language LoRA parameters. 
%\priyanka{Should elaborate why} \alex{I thought this is answered in the next paragraph}
We highlight that the language adapters are trained by fine-tuning \LLM{} with LoRA on prefix-LM for just $5k$ steps; even with this minimal training, they provide knowledge that is helpful to the pretrained model.

\noindent\textbf{Why is subtracting the source language adapter important?} We hypothesize that since the task adapter encodes information on summarizing articles in En (source), it is beneficial to add a language adapter that encourages the LLM to generate in the target language, but at the same time avoid generating in the source. Intuitively, negating the En language adapter parameters likely reduces the bias of the model towards En and enhances the ability of the model to generate in the target language.

% \noindent \textbf{Oracle approaches.} 
% As an oracle experiment, we train a task adapter using LoRA parameters in target language data (individually for each language in $\text{XLSum}_{unseen}$). On average, performance improves by 3.5 ROUGE points compared to the task adapter trained on English labeled data (first row). In the fifth row (\textit{full fine-tuning}), we show the results we get by fully fine-tuning \LLM{} (i.e., fine-tuning all the parameters of the model) on target labeled data of each target language. \LLM{} fine-tuning only achieves +0.6 ROUGE points compared to the task adapter oracle approach. This result further motivates the use of LoRA layers or other PEFT methods when adapting multilingual LLMs, such as \LLM{}.

\subsection{Task-in-Many-Languages}

We present the results of our approach when task data is available in different languages in Table \ref{table:taskinmanylangs}. We compare the baselines with  \textit{task-only; Add all}, which fine-tunes \LLM{} with LoRA on each language of the training set, and then computes the weight average of all fine-tuned models. 

\noindent\textbf{Task-only (Add all) on par with multilingual baseline:} 
% We note that the \textit{baseline (best)}, the \textit{baseline (multilingual)} and the \textit{task addition (all)} approaches perform equivalently. This means that averaging \textit{weights} of fine-tuned models is in this case on par with multilingually fine-tuning a LLM for a given task. 
We observe that simply averaging all task adapters is on par with the multilingual baseline. This is intriguing, as it suggests that model merging can be used to iteratively add new task data to a petrained model. As soon as new task data (for a previously unsupported language) become available, one can simply train the corresponding task vector on this data and add it to the model by performing weight averaging. This alleviates the need of training a new multilingual model for every new batch of data.

\noindent\textbf{Adding only related task adapters gives better results for most languages:} 
Our approach \textit{(task-only; Add related)} is presented in row 4. This selective composition of task adapters clearly surpasses the baselines. Our hypothesis is that not all task adapters are as important for a target language $T$ and the final model should only incorporate task adapters trained in languages similar to the target. To select the models that will be averaged, we do not use any test data, but rely on linguistic information.  We query the URIEL database and use the languages with the smallest distance to each held-out language $T$. Our approach outperforms the uniform weight average \textit{(task-only; Add all)}, likely because our model avoids negative transfer between task adapters learned on distant languages, and leverages task information learned from similar languages. 

\noindent\textbf{Arithmetically composing language and task adapters when task data is available in multiple languages is not helpful:} 
We present the results we computed using \textit{Language and Task; Add and Subtract related} which leverages unlabeled data as well as task data in the final row of Table \ref{table:taskinmanylangs}. This approach performs on par with the \textit{task-only; Add related} approach that uses only labeled data. Composing language and task knowledge is beneficial in the absence of enough task data. However, when task data is available in multiple languages, combining information from similar languages yields strong results and unlabeled data does not provide an additional benefit. Therefore, merging the two methods does not provide improvements. 

\section{Analysis}
\subsection{Using task adapter in different languages has consistent improvements}
\begin{figure}[t]
	\centering
	\includegraphics[width=0.85\columnwidth, page=1]{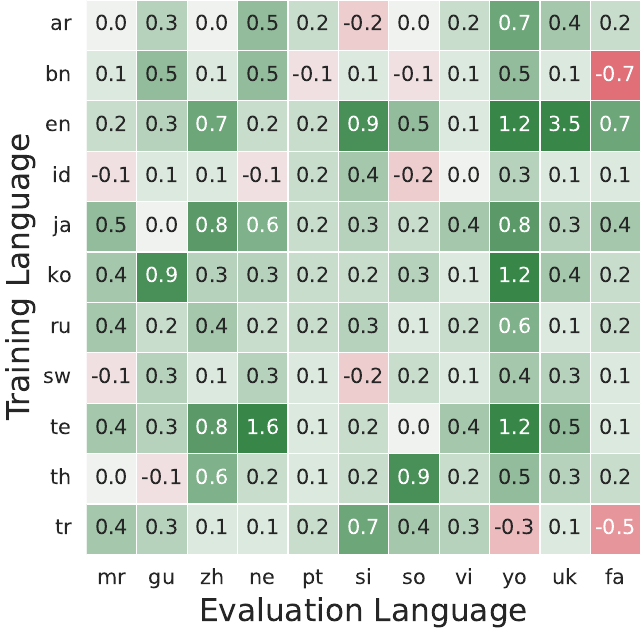}
        	\caption{Relative ROUGE-2 improvement of our \textbf{language \& task arithmetic} over the baseline (task adapter only). Our approach yields consistent improvements for most source-target language pairs.}
         \label{fig:arithmeticheatmap}
\end{figure}

For our main language and task arithmetic results with \textit{Task-in-One-Language}, we trained the task adapter on En labeled data and evaluated its performance on $\text{XLSum}_{unseen}$. For a more fine-grained assessment of our model, we present its relative performance when the task adapter is trained in each language in $\text{XLSum}_{seen}$ (as opposed to just En) against the corresponding baseline. The results are shown in Figure \ref{fig:arithmeticheatmap}. The third row ({En}) shows the performance difference of \textit{Language and Task (Add and Subtract)} from the baseline (Table \ref{table:arithmeticresults}). 

We observe consistent improvements using our approach compared to the baseline across all language pairs. Low-resource languages, 
such as Yo, benefit more from the cross-lingual transfer setup we propose. In addition, while learning the En task adapter seems to provide higher gains for most evaluation languages, Te, Ja and Ko task adapters also lead to a large performance boost. 

% \paragraph{Adding Language-Specific Capacity Boosts Zero-Shot  Transfer:}
While \LLM{} has been trained on vast multilingual data, providing each language with individual capacity using language modeling yields across-the-board improvements. This suggests that learning language-specific knowledge using PEFT parameters has the potential to strengthen the zero-shot cross-lingual transfer abilities of LLMs at a very small computational cost. 
% Based on our experiments across multiple languages, our method can efficiently and effectively strengthen the zero-shot capabilities of multilingual LLMs. 

\begin{table*}[t]
\resizebox{0.95\textwidth}{!}{
\begin{tabular}{lrrrrrrrrrrrr}
\toprule
 \textbf{Method}  & Mr & Gu & Zh & Ne & Pt & Si & So & Vi & Yo & Uk & Fa &  \multicolumn{1}{c}{Avg} \\
\midrule 
 \multicolumn{5}{l}{\textbf{Task-in-Many-Languages}} & & &&&& & \\ 
Baseline (best) & 21.3 & 31.4 & 25.6 & 30.0 & 22.6 & 36.0 & 22.9 & 25.4 & 21.8 & 22.0 & 25.7 & 25.9 \\
Baseline (multilingual) & 21.2 & 31.5 &  26.1 & 30.8  & 23.2 & 36.7 & 23.1 & 25.5 & 21.5 & 22.0 & 25.9 & 26.1 \\ \midrule 
Task-only (Add all) & 20.9 & 31.3  &  25.6 & 30.5 & 22.8  &35.9  & 22.7 & 25.2 & 20.8 & 21.9 & 25.7 &  25.7 \\ 
Task-only (Add related) &  21.1 &  \textbf{32.2} & 26.2 & \textbf{31.4} & \textbf{24.0} & 36.6 & 22.9 & \textbf{25.7} & 21.9 & \textbf{22.3} & \textbf{26.6} & \textbf{26.4} \\ 
% \midrule 
% \multicolumn{7}{l}{\textbf{\textbf{Oracle approaches (using target language labeled data)}}} &  &&&& & \\ 
% & LoRA  & 24.2 & 32.5 &27.5 &  31.5& 24.4 & 36.3 & 23.5 & 26.2 & 28.3 & 22.9 & 27.4 & 27.7 \\ 
% & Kronecker  & 24.3 & 32.8  & 27.8  & 31.9 & 24.2 & 35.0 & 24.0 & 26.4 & 28.1 & 23.2 & 27.7 & 27.8 \\ 
\bottomrule   
\end{tabular}
} \caption{\textbf{Adding related task adapters outperforms monolingual and multilingual baselines on XLSum using Kronecker adapter.} Rouge (ROUGE-2 spm) zero-shot scores on the $\text{XLSum}_{unseen}$ test set.}
\label{table:kron}
\end{table*}

%\subsection{Comparison of LoRA to other PEFT methods}
\subsection{Our method also works with other PEFT parameters}
We showed that composing task and language LoRA weights by element-wise arithmetic brings significant gains to cross-lingual transfer. In this section, we examine whether our findings also generalize to parameter-efficient fine-tuning methods other than LoRA.
 
 One particularly interesting PEFT method is Kronecker adapter \cite{edalati2022krona}. While LoRA is based on the multiplication of two low-rank matrices, Kronecker adapter is a matrix decomposition method which does not rely on the low-rank assumption. Instead, it replaces the low-rank decomposition in LoRA with the Kronecker product decomposition. It has been shown that this PEFT method achieves large improvements over LoRA and full fine-tuning 
 on the GLUE benchmark~\cite{wang-etal-2018-glue}. We conduct language and task arithmetic using Kronecker adapters 
  as the PEFT modules.\footnote{Similar to LoRA tuning, we add Kronecker adapters for the Key, Query, Value, Projection attention matrices of the Transformer model while keeping the weights fixed.} 

\noindent\textbf{Kronecker adapter:}
Formally, the Kronecker product is defined as follows:
\begin{equation*}
A \otimes B = 
\begin{pmatrix}
a_{11}B & \cdots & a_{1n}B \\
\vdots  & \ddots & \vdots  \\
a_{m1}B & \cdots & a_{mn}B 
\end{pmatrix}
\end{equation*}
where matrices $\mathbf{A} \in \real^{m \times n}$ and $\mathbf{B} \in \real^{ \frac{k}{m} \times \frac{d}{n}}$ are the input matrices, and $\mathbf{W} \in \real^{k \times d}$, $k$ is the model dimension and $d$ is the dimension per attention head is the output matrix. We can tune hyperparameters $m$ and $n$ while keeping the number of additional parameters fixed, which is more flexible than LoRA. 

\noindent\textbf{Experimental setting:} We use \LLM{} S model as the pretrained LLM. We add a Kronecker adapter with $(m, n)=(32, 16)$. Similar to LoRA, this PEFT method does not decrease inference speed because the additional parameters are added back to the original model weights.

\noindent\textbf{Results:} We run the \textit{task-only; Add} experiments using Kronecker adapter and show the results in Table \ref{table:kron}. 
We observe that the results follow a similar pattern as with the LoRA adapter. Our method (\textit{task-only; Add related}) outperforms  monolingual and multilingual baselines. This demonstrates that a selective combination of PEFT parameters at the weight level improves the generalization ability of a LLM to languages for which no task data is available. This confirms our intuition that it is possible to compose information learned about a task in different languages by simply performing point-wise operations.

\subsection{Module subtraction is particularly helpful for summarization}
We proposed two composition approaches for language and task arithmetic: \textit{Add} or \textit{Add and Subtract}.
To understand the different impact of these two approaches, we compare their performance on two datasets, TyDi QA and XLSum.

\noindent\textbf{Experimental setting}:
Besides XLSum, we also evaluate our language and task arithmetic approach on TyDi QA \cite{clark-etal-2020-tydi}, a multilingual extractive question answering dataset of 8 typologically diverse languages, based on Wikipedia articles in Bengali ({bn}), English ({en}), Finnish ({fi}), Indonesian ({id}), Korean ({ko}), Russian ({ru}), Swahili ({sw}), and Telugu ({te}). We train our model on En task data an evaluate on each of the other languages in the dataset, simulating a zero-shot setup.

\noindent\textbf{Results}:
We show the results in Table \ref{table:ablation}. We find that using both addition and subtraction is more beneficial than addition only for XLSum~($+0.6$ gains in ROUGE). However, we observe that for the QA task, using addition and subtraction performs on par with addition only. We hypothesize that this is likely because TyDi QA is an extractive QA task where the model simply needs to copy a segment of correct answer from the context, while XLSum requires more free-form language generation. Because of this inherent difference between the tasks, discouraging the model from generating in the source language (by negating the source language adapter) is less essential to QA compared to summarization.

\begin{table}[h]
\resizebox{0.90\columnwidth}{!}{
\begin{tabular}{lcc}
\toprule
  \textbf{Method}  &  TyDi QA & XLSum \\
  \midrule 
Baseline & 83.0 & 24.2 \\
\multicolumn{3}{l}{Language and task arithmetic} \\
\hspace{4mm} - Add & 83.3 & 24.4 \\ 
\hspace{4mm} - Add and Subtract & 83.2 & 25.0 \\ 
 \bottomrule 

\end{tabular} \label{table:debiasing}

} \caption{\textbf{Language and task arithmetic via addition or addition and subtraction for TyDi QA and XLSum} using LoRA parameters. These are the average results over the unseen languages. For TyDi QA, F1 is shown, while for XLSum, we show ROUGE-2 spm.}
\label{table:ablation}
\end{table}

\subsection{Task adapters selected by \textit{lang2vec}}
\label{subsection:l2v}
When we have labeled data available in multiple languages, our proposed \textit{task-only; Add related} approach averages the weights of PEFT parameters that are related to the target language.  The relatedness is defined by \textit{lang2vec}, a tool that queries URIEL. To shed light on where the improved performance of our model comes from, we present in Table \ref{table:l2v} the source languages that are selected for each of the target languages based on linguistic knowledge.

We witness that a different number of languages is selected for each target language. We do not explicitly control the number of models averaged, we simply sort them using the syntactic and geographic distance. % as quantified in URIEL.
For a given target language $T$, we average the weights of the source languages $S_1, S_2, .., S_N$ that have a syntactic distance < 0.7 and a geographic distance < 0.3. 
We leave a more fine-grained selection process to future work. 

\begin{table}[]
\resizebox{0.95\columnwidth}{!}{
\begin{tabular}{rrrrrrrrrrr}
\toprule
 Mr & Gu & Zh & Ne & Pt & Si & So & Vi & Yo & Uk & Fa \\ \midrule 
Bn & Bn & En & Te & En & Te & Ar & Id & En & Ru & Tr \\
 Te & Te   & Ko & Ja & Ru & Bn & Sw & Th & Ar & En & En \\ 
 Tr &      & Ja & Tr & Ar &    & En  & & & Sw & Ar \\ 
  &        & Id & Ko &    & & & & & & \\ 
  &        & Th & Ru &    & & & & & & \\ 
  &        &  & Bn &    & & & & & & \\ 

\bottomrule   
\end{tabular}

} \caption{Most similar languages to each of the evaluation languages (based on lang2vec) selected by our \textit{task-only (Add related)} approach.}\label{table:l2v}
\end{table}

\section{Related Work}

LLMs have shown impressive performance in various natural language processing tasks \cite{radford2019language, brown, chung2022scaling, touvron2023llama}, often requiring no extra training to adapt to downstream tasks.

Numerous parameter-efficient methods have been proposed, each addressing the challenge of enhancing efficiency
% in various ways
. These methods can be categorized as input composition, function composition, and parameter composition \cite{pfeiffer2023modulardeeplearning}. \textit{Input composition} methods, such as prompt tuning, incorporate soft prompts into the input layers to guide the model's behavior~\citep{li-liang-2021-prefix, lester-etal-2021-power}.  \textit{Function composition} strategies, like  adapters~\citep{rebuffi, houlsby}, introduce non-linear functions within pretrained layers to adapt the intermediate representations of the model. 
% Despite their effectiveness in outperforming prompt-based adaptations, they  tend to require a higher number of trainable parameters. 
\textit{Parameter composition} is exemplified by methods like LoRA~\cite{hu2022lora}, which introduces a limited number of learnable low-rank matrices into each pretrained layer.

Recent work which is based on the linear mode connectivity \cite{frankle2020} suggests averaging the weights of pretrained models fine-tuned on the same dataset with different hyperparameters to improve downstream performance \cite{izmailov2018averaging, stochastic-weight-averaging-in-parallel, pmlr-v162-wortsman22a}. It has also been shown that averaging the weights of models fine-tuned on different tasks improves out-of-domain generalization without leaking information about potentially private labeled datasets \cite{jin2023dataless}. Composing weights of models fine-tuned on tasks related to the target task is also beneficial \cite{fisheravg}. \citet{ainsworth2023git, ilharco2023editing, yadav2023tiesmerging, huang2023lorahub,ortizjimenez2023task} show that a model can acquire multi-task learning abilities using model merging, while \citet{daheim2024model} propose model merging by reducing gradient mismatch. 
There is also work on averaging domain-specific adapter layers \cite{chronopoulou-etal-2023-adaptersoup} or domain-expert LMs \cite{btm-suchin} with large gains for unseen domains. However, there is no work on PEFT cross-lingual transfer using language and task arithmetic. 

In a similar line of thought and to mitigate interference of different tasks during training, \citet{pfeiffer-etal-2021-adapterfusion}  train task PEFT modules and learn attention parameters to select the most useful of them, while \citet{karimi-mahabadi-etal-2021-parameter} learn adapters with hypernetworks.  \citet{asai-etal-2022-attempt} efficiently integrate knowledge from multiple tasks with a mix of trainable soft prompts. \citet{ponti-etal-2023-combining} propose Polytropon, which learns both adapters and a binary task–module routing matrix, determining which module should be active for each task; \citet{caccia2023multihead} extend it to a more granular level by mixing subsets of adapter dimensions. 

Another research direction considers training  PEFT parameters and combining them for cross-lingual transfer. MAD-X \cite{pfeiffer-etal-2020-mad} stacks task bottleneck adapters with language adapters and using them for cross-lingual transfer. \citet{ansell-etal-2022-composable}  identify the parameters that are most useful for a task and a language, and compose them; this work is  based on the lottery ticket hypothesis \cite{frankle2020}. \citet{vu-etal-2022-overcoming} propose factorizing a prompt into a language and task and training each part while keeping the other frozen. Newly learned knowledge is combined with the existing model using PEFT modules to permit cross-lingual transfer in multiple recent works \cite{bapna-firat-2019-simple, ustun-etal-2020-udapter, vidoni2020orthogonal, cooper-stickland-etal-2021-recipes, chronopoulou-etal-2023-language}. To the best of our knowledge, our work is the first to propose improving cross-lingual transfer of a LLM via a combination of weights of PEFT parameters.

\section{Conclusion}
We present a new method to compose knowledge from parameter-efficient modules using arithmetic operations in order to improve zero-shot cross-lingual transfer. Our experiments in summarization on a wide set of languages using \LLM{} as the pretrained model show that our \textit{language and task arithmetic} achieves consistent improvements over the baselines and introduces a modular approach that can be leveraged for improved generalization of a LLM in languages that lack labeled data.

\bibliography{anthology,custom}

\begin{thebibliography}{61}
\expandafter\ifx\csname natexlab\endcsname\relax\def\natexlab#1{#1}\fi

\bibitem[{Ainsworth et~al.(2023)Ainsworth, Hayase, and
  Srinivasa}]{ainsworth2023git}
Samuel Ainsworth, Jonathan Hayase, and Siddhartha Srinivasa. 2023.
\newblock \href {https://openreview.net/forum?id=CQsmMYmlP5T} {Git re-basin:
  Merging models modulo permutation symmetries}.
\newblock In \emph{The Eleventh International Conference on Learning
  Representations}.

\bibitem[{Anil et~al.(2023)Anil, Dai, Firat, Johnson, Lepikhin, Passos,
  Shakeri, Taropa, Bailey, Chen, Chu, Clark, Shafey, Huang, Meier-Hellstern,
  Mishra, Moreira, Omernick, Robinson, Ruder, Tay, Xiao, Xu, Zhang, Abrego,
  Ahn, Austin, Barham, Botha, Bradbury, Brahma, Brooks, Catasta, Cheng, Cherry,
  Choquette-Choo, Chowdhery, Crepy, Dave, Dehghani, Dev, Devlin, Díaz, Du,
  Dyer, Feinberg, Feng, Fienber, Freitag, Garcia, Gehrmann, Gonzalez, Gur-Ari,
  Hand, Hashemi, Hou, Howland, Hu, Hui, Hurwitz, Isard, Ittycheriah, Jagielski,
  Jia, Kenealy, Krikun, Kudugunta, Lan, Lee, Lee, Li, Li, Li, Li, Li, Lim, Lin,
  Liu, Liu, Maggioni, Mahendru, Maynez, Misra, Moussalem, Nado, Nham, Ni,
  Nystrom, Parrish, Pellat, Polacek, Polozov, Pope, Qiao, Reif, Richter, Riley,
  Ros, Roy, Saeta, Samuel, Shelby, Slone, Smilkov, So, Sohn, Tokumine, Valter,
  Vasudevan, Vodrahalli, Wang, Wang, Wang, Wang, Wieting, Wu, Xu, Xu, Xue, Yin,
  Yu, Zhang, Zheng, Zheng, Zhou, Zhou, Petrov, and Wu}]{palm2}
Rohan Anil, Andrew~M. Dai, Orhan Firat, Melvin Johnson, Dmitry Lepikhin,
  Alexandre Passos, Siamak Shakeri, Emanuel Taropa, Paige Bailey, Zhifeng Chen,
  Eric Chu, Jonathan~H. Clark, Laurent~El Shafey, Yanping Huang, Kathy
  Meier-Hellstern, Gaurav Mishra, Erica Moreira, Mark Omernick, Kevin Robinson,
  Sebastian Ruder, Yi~Tay, Kefan Xiao, Yuanzhong Xu, Yujing Zhang,
  Gustavo~Hernandez Abrego, Junwhan Ahn, Jacob Austin, Paul Barham, Jan Botha,
  James Bradbury, Siddhartha Brahma, Kevin Brooks, Michele Catasta, Yong Cheng,
  Colin Cherry, Christopher~A. Choquette-Choo, Aakanksha Chowdhery, Clément
  Crepy, Shachi Dave, Mostafa Dehghani, Sunipa Dev, Jacob Devlin, Mark Díaz,
  Nan Du, Ethan Dyer, Vlad Feinberg, Fangxiaoyu Feng, Vlad Fienber, Markus
  Freitag, Xavier Garcia, Sebastian Gehrmann, Lucas Gonzalez, Guy Gur-Ari,
  Steven Hand, Hadi Hashemi, Le~Hou, Joshua Howland, Andrea Hu, Jeffrey Hui,
  Jeremy Hurwitz, Michael Isard, Abe Ittycheriah, Matthew Jagielski, Wenhao
  Jia, Kathleen Kenealy, Maxim Krikun, Sneha Kudugunta, Chang Lan, Katherine
  Lee, Benjamin Lee, Eric Li, Music Li, Wei Li, YaGuang Li, Jian Li, Hyeontaek
  Lim, Hanzhao Lin, Zhongtao Liu, Frederick Liu, Marcello Maggioni, Aroma
  Mahendru, Joshua Maynez, Vedant Misra, Maysam Moussalem, Zachary Nado, John
  Nham, Eric Ni, Andrew Nystrom, Alicia Parrish, Marie Pellat, Martin Polacek,
  Alex Polozov, Reiner Pope, Siyuan Qiao, Emily Reif, Bryan Richter, Parker
  Riley, Alex~Castro Ros, Aurko Roy, Brennan Saeta, Rajkumar Samuel, Renee
  Shelby, Ambrose Slone, Daniel Smilkov, David~R. So, Daniel Sohn, Simon
  Tokumine, Dasha Valter, Vijay Vasudevan, Kiran Vodrahalli, Xuezhi Wang,
  Pidong Wang, Zirui Wang, Tao Wang, John Wieting, Yuhuai Wu, Kelvin Xu, Yunhan
  Xu, Linting Xue, Pengcheng Yin, Jiahui Yu, Qiao Zhang, Steven Zheng,
  Ce~Zheng, Weikang Zhou, Denny Zhou, Slav Petrov, and Yonghui Wu. 2023.
\newblock \href {http://arxiv.org/abs/2305.10403} {Palm 2 technical report}.

\bibitem[{Ansell et~al.(2022)Ansell, Ponti, Korhonen, and
  Vuli{\'c}}]{ansell-etal-2022-composable}
Alan Ansell, Edoardo Ponti, Anna Korhonen, and Ivan Vuli{\'c}. 2022.
\newblock \href {https://doi.org/10.18653/v1/2022.acl-long.125} {Composable
  sparse fine-tuning for cross-lingual transfer}.
\newblock In \emph{Proceedings of the 60th Annual Meeting of the Association
  for Computational Linguistics (Volume 1: Long Papers)}, pages 1778--1796,
  Dublin, Ireland. Association for Computational Linguistics.

\bibitem[{Asai et~al.(2022)Asai, Salehi, Peters, and
  Hajishirzi}]{asai-etal-2022-attempt}
Akari Asai, Mohammadreza Salehi, Matthew Peters, and Hannaneh Hajishirzi. 2022.
\newblock \href {https://doi.org/10.18653/v1/2022.emnlp-main.446} {{ATTEMPT}:
  Parameter-efficient multi-task tuning via attentional mixtures of soft
  prompts}.
\newblock In \emph{Proceedings of the 2022 Conference on Empirical Methods in
  Natural Language Processing}, pages 6655--6672, Abu Dhabi, United Arab
  Emirates. Association for Computational Linguistics.

\bibitem[{Bapna and Firat(2019)}]{bapna-firat-2019-simple}
Ankur Bapna and Orhan Firat. 2019.
\newblock \href {https://doi.org/10.18653/v1/D19-1165} {Simple, scalable
  adaptation for neural machine translation}.
\newblock In \emph{Proceedings of the 2019 Conference on Empirical Methods in
  Natural Language Processing and the 9th International Joint Conference on
  Natural Language Processing (EMNLP-IJCNLP)}, pages 1538--1548, Hong Kong,
  China. Association for Computational Linguistics.

\bibitem[{Brown et~al.(2020)Brown, Mann, Ryder, Subbiah, Kaplan, Dhariwal,
  Neelakantan, Shyam, Sastry, Askell, Agarwal, Herbert-Voss, Krueger, Henighan,
  Child, Ramesh, Ziegler, Wu, Winter, Hesse, Chen, Sigler, Litwin, Gray, Chess,
  Clark, Berner, McCandlish, Radford, Sutskever, and Amodei}]{brown}
Tom Brown, Benjamin Mann, Nick Ryder, Melanie Subbiah, Jared~D Kaplan, Prafulla
  Dhariwal, Arvind Neelakantan, Pranav Shyam, Girish Sastry, Amanda Askell,
  Sandhini Agarwal, Ariel Herbert-Voss, Gretchen Krueger, Tom Henighan, Rewon
  Child, Aditya Ramesh, Daniel Ziegler, Jeffrey Wu, Clemens Winter, Chris
  Hesse, Mark Chen, Eric Sigler, Mateusz Litwin, Scott Gray, Benjamin Chess,
  Jack Clark, Christopher Berner, Sam McCandlish, Alec Radford, Ilya Sutskever,
  and Dario Amodei. 2020.
\newblock \href
  {https://proceedings.neurips.cc/paper_files/paper/2020/file/1457c0d6bfcb4967418bfb8ac142f64a-Paper.pdf}
  {Language models are few-shot learners}.
\newblock In \emph{Advances in Neural Information Processing Systems},
  volume~33, pages 1877--1901. Curran Associates, Inc.

\bibitem[{Caccia et~al.(2023)Caccia, Ponti, Su, Pereira, Roux, and
  Sordoni}]{caccia2023multihead}
Lucas Caccia, Edoardo Ponti, Zhan Su, Matheus Pereira, Nicolas~Le Roux, and
  Alessandro Sordoni. 2023.
\newblock \href {http://arxiv.org/abs/2211.03831} {Multi-head adapter routing
  for cross-task generalization}.

\bibitem[{Cao et~al.(2020)Cao, Wan, Yao, and Yu}]{Cao_Wan_Yao_Yu_2020}
Yue Cao, Xiaojun Wan, Jinge Yao, and Dian Yu. 2020.
\newblock \href {https://doi.org/10.1609/aaai.v34i01.5328} {Multisumm: Towards
  a unified model for multi-lingual abstractive summarization}.
\newblock \emph{Proceedings of the AAAI Conference on Artificial Intelligence},
  34(01):11--18.

\bibitem[{Chowdhery et~al.(2022)Chowdhery, Narang, Devlin, Bosma, Mishra,
  Roberts, Barham, Chung, Sutton, Gehrmann, Schuh, Shi, Tsvyashchenko, Maynez,
  Rao, Barnes, Tay, Shazeer, Prabhakaran, Reif, Du, Hutchinson, Pope, Bradbury,
  Austin, Isard, Gur-Ari, Yin, Duke, Levskaya, Ghemawat, Dev, Michalewski,
  Garcia, Misra, Robinson, Fedus, Zhou, Ippolito, Luan, Lim, Zoph, Spiridonov,
  Sepassi, Dohan, Agrawal, Omernick, Dai, Pillai, Pellat, Lewkowycz, Moreira,
  Child, Polozov, Lee, Zhou, Wang, Saeta, Diaz, Firat, Catasta, Wei,
  Meier-Hellstern, Eck, Dean, Petrov, and Fiedel}]{chowdhery2022palm}
Aakanksha Chowdhery, Sharan Narang, Jacob Devlin, Maarten Bosma, Gaurav Mishra,
  Adam Roberts, Paul Barham, Hyung~Won Chung, Charles Sutton, Sebastian
  Gehrmann, Parker Schuh, Kensen Shi, Sasha Tsvyashchenko, Joshua Maynez,
  Abhishek Rao, Parker Barnes, Yi~Tay, Noam Shazeer, Vinodkumar Prabhakaran,
  Emily Reif, Nan Du, Ben Hutchinson, Reiner Pope, James Bradbury, Jacob
  Austin, Michael Isard, Guy Gur-Ari, Pengcheng Yin, Toju Duke, Anselm
  Levskaya, Sanjay Ghemawat, Sunipa Dev, Henryk Michalewski, Xavier Garcia,
  Vedant Misra, Kevin Robinson, Liam Fedus, Denny Zhou, Daphne Ippolito, David
  Luan, Hyeontaek Lim, Barret Zoph, Alexander Spiridonov, Ryan Sepassi, David
  Dohan, Shivani Agrawal, Mark Omernick, Andrew~M. Dai,
  Thanumalayan~Sankaranarayana Pillai, Marie Pellat, Aitor Lewkowycz, Erica
  Moreira, Rewon Child, Oleksandr Polozov, Katherine Lee, Zongwei Zhou, Xuezhi
  Wang, Brennan Saeta, Mark Diaz, Orhan Firat, Michele Catasta, Jason Wei,
  Kathy Meier-Hellstern, Douglas Eck, Jeff Dean, Slav Petrov, and Noah Fiedel.
  2022.
\newblock \href {http://arxiv.org/abs/2204.02311} {Palm: Scaling language
  modeling with pathways}.

\bibitem[{Chronopoulou et~al.(2023{\natexlab{a}})Chronopoulou, Peters, Fraser,
  and Dodge}]{chronopoulou-etal-2023-adaptersoup}
Alexandra Chronopoulou, Matthew Peters, Alexander Fraser, and Jesse Dodge.
  2023{\natexlab{a}}.
\newblock \href {https://aclanthology.org/2023.findings-eacl.153}
  {{A}dapter{S}oup: Weight averaging to improve generalization of pretrained
  language models}.
\newblock In \emph{Findings of the Association for Computational Linguistics:
  EACL 2023}, pages 2054--2063, Dubrovnik, Croatia. Association for
  Computational Linguistics.

\bibitem[{Chronopoulou et~al.(2023{\natexlab{b}})Chronopoulou, Stojanovski, and
  Fraser}]{chronopoulou-etal-2023-language}
Alexandra Chronopoulou, Dario Stojanovski, and Alexander Fraser.
  2023{\natexlab{b}}.
\newblock \href {https://doi.org/10.18653/v1/2023.loresmt-1.5} {Language-family
  adapters for low-resource multilingual neural machine translation}.
\newblock In \emph{Proceedings of the The Sixth Workshop on Technologies for
  Machine Translation of Low-Resource Languages (LoResMT 2023)}, pages 59--72,
  Dubrovnik, Croatia. Association for Computational Linguistics.

\bibitem[{Chung et~al.(2022)Chung, Hou, Longpre, Zoph, Tay, Fedus, Li, Wang,
  Dehghani, Brahma, Webson, Gu, Dai, Suzgun, Chen, Chowdhery, Castro-Ros,
  Pellat, Robinson, Valter, Narang, Mishra, Yu, Zhao, Huang, Dai, Yu, Petrov,
  Chi, Dean, Devlin, Roberts, Zhou, Le, and Wei}]{chung2022scaling}
Hyung~Won Chung, Le~Hou, Shayne Longpre, Barret Zoph, Yi~Tay, William Fedus,
  Yunxuan Li, Xuezhi Wang, Mostafa Dehghani, Siddhartha Brahma, Albert Webson,
  Shixiang~Shane Gu, Zhuyun Dai, Mirac Suzgun, Xinyun Chen, Aakanksha
  Chowdhery, Alex Castro-Ros, Marie Pellat, Kevin Robinson, Dasha Valter,
  Sharan Narang, Gaurav Mishra, Adams Yu, Vincent Zhao, Yanping Huang, Andrew
  Dai, Hongkun Yu, Slav Petrov, Ed~H. Chi, Jeff Dean, Jacob Devlin, Adam
  Roberts, Denny Zhou, Quoc~V. Le, and Jason Wei. 2022.
\newblock \href {http://arxiv.org/abs/2210.11416} {Scaling
  instruction-finetuned language models}.

\bibitem[{Clark et~al.(2020)Clark, Choi, Collins, Garrette, Kwiatkowski,
  Nikolaev, and Palomaki}]{clark-etal-2020-tydi}
Jonathan~H. Clark, Eunsol Choi, Michael Collins, Dan Garrette, Tom Kwiatkowski,
  Vitaly Nikolaev, and Jennimaria Palomaki. 2020.
\newblock \href {https://doi.org/10.1162/tacl_a_00317} {{T}y{D}i {QA}: A
  benchmark for information-seeking question answering in typologically diverse
  languages}.
\newblock \emph{Transactions of the Association for Computational Linguistics},
  8:454--470.

\bibitem[{Cooper~Stickland et~al.(2021)Cooper~Stickland, Li, and
  Ghazvininejad}]{cooper-stickland-etal-2021-recipes}
Asa Cooper~Stickland, Xian Li, and Marjan Ghazvininejad. 2021.
\newblock \href {https://doi.org/10.18653/v1/2021.eacl-main.301} {Recipes for
  adapting pre-trained monolingual and multilingual models to machine
  translation}.
\newblock In \emph{Proceedings of the 16th Conference of the European Chapter
  of the Association for Computational Linguistics: Main Volume}, pages
  3440--3453, Online. Association for Computational Linguistics.

\bibitem[{Daheim et~al.(2024)Daheim, M{\"o}llenhoff, Ponti, Gurevych, and
  Khan}]{daheim2024model}
Nico Daheim, Thomas M{\"o}llenhoff, Edoardo Ponti, Iryna Gurevych, and
  Mohammad~Emtiyaz Khan. 2024.
\newblock \href {https://openreview.net/forum?id=D7KJmfEDQP} {Model merging by
  uncertainty-based gradient matching}.
\newblock In \emph{The Twelfth International Conference on Learning
  Representations}.

\bibitem[{Edalati et~al.(2022)Edalati, Tahaei, Kobyzev, Nia, Clark, and
  Rezagholizadeh}]{edalati2022krona}
Ali Edalati, Marzieh Tahaei, Ivan Kobyzev, Vahid~Partovi Nia, James~J. Clark,
  and Mehdi Rezagholizadeh. 2022.
\newblock \href {http://arxiv.org/abs/2212.10650} {Krona: Parameter efficient
  tuning with kronecker adapter}.

\bibitem[{Fifty et~al.(2021)Fifty, Amid, Zhao, Yu, Anil, and
  Finn}]{NEURIPS2021_e77910eb}
Chris Fifty, Ehsan Amid, Zhe Zhao, Tianhe Yu, Rohan Anil, and Chelsea Finn.
  2021.
\newblock \href
  {https://proceedings.neurips.cc/paper_files/paper/2021/file/e77910ebb93b511588557806310f78f1-Paper.pdf}
  {Efficiently identifying task groupings for multi-task learning}.
\newblock In \emph{Advances in Neural Information Processing Systems},
  volume~34, pages 27503--27516. Curran Associates, Inc.

\bibitem[{Frankle and Carbin(2019)}]{frankle2018the}
Jonathan Frankle and Michael Carbin. 2019.
\newblock \href {https://openreview.net/forum?id=rJl-b3RcF7} {The lottery
  ticket hypothesis: Finding sparse, trainable neural networks}.
\newblock In \emph{International Conference on Learning Representations}.

\bibitem[{Frankle et~al.(2020)Frankle, Dziugaite, Roy, and
  Carbin}]{frankle2020}
Jonathan Frankle, Gintare~Karolina Dziugaite, Daniel Roy, and Michael Carbin.
  2020.
\newblock \href {https://proceedings.mlr.press/v119/frankle20a.html} {Linear
  mode connectivity and the lottery ticket hypothesis}.
\newblock In \emph{Proceedings of the 37th International Conference on Machine
  Learning}, volume 119 of \emph{Proceedings of Machine Learning Research},
  pages 3259--3269. PMLR.

\bibitem[{Giannakopoulos et~al.(2015)Giannakopoulos, Kubina, Conroy,
  Steinberger, Favre, Kabadjov, Kruschwitz, and
  Poesio}]{giannakopoulos-etal-2015-multiling}
George Giannakopoulos, Jeff Kubina, John Conroy, Josef Steinberger, Benoit
  Favre, Mijail Kabadjov, Udo Kruschwitz, and Massimo Poesio. 2015.
\newblock \href {https://doi.org/10.18653/v1/W15-4638} {{M}ulti{L}ing 2015:
  Multilingual summarization of single and multi-documents, on-line fora, and
  call-center conversations}.
\newblock In \emph{Proceedings of the 16th Annual Meeting of the Special
  Interest Group on Discourse and Dialogue}, pages 270--274, Prague, Czech
  Republic. Association for Computational Linguistics.

\bibitem[{Grusky et~al.(2018)Grusky, Naaman, and
  Artzi}]{grusky-etal-2018-newsroom}
Max Grusky, Mor Naaman, and Yoav Artzi. 2018.
\newblock \href {https://doi.org/10.18653/v1/N18-1065} {{N}ewsroom: A dataset
  of 1.3 million summaries with diverse extractive strategies}.
\newblock In \emph{Proceedings of the 2018 Conference of the North {A}merican
  Chapter of the Association for Computational Linguistics: Human Language
  Technologies, Volume 1 (Long Papers)}, pages 708--719, New Orleans,
  Louisiana. Association for Computational Linguistics.

\bibitem[{Gupta et~al.(2020)Gupta, Serrano, and
  DeCoste}]{stochastic-weight-averaging-in-parallel}
Vipul Gupta, Santiago~Akle Serrano, and Dennis DeCoste. 2020.
\newblock \href {https://arxiv.org/pdf/2001.02312.pdf} {Stochastic weight
  averaging in parallel: Large-batch training that generalizes well}.
\newblock In \emph{ICLR}.

\bibitem[{Hasan et~al.(2021)Hasan, Bhattacharjee, Islam, Mubasshir, Li, Kang,
  Rahman, and Shahriyar}]{hasan-etal-2021-xl}
Tahmid Hasan, Abhik Bhattacharjee, Md.~Saiful Islam, Kazi Mubasshir, Yuan-Fang
  Li, Yong-Bin Kang, M.~Sohel Rahman, and Rifat Shahriyar. 2021.
\newblock \href {https://doi.org/10.18653/v1/2021.findings-acl.413} {{XL}-sum:
  Large-scale multilingual abstractive summarization for 44 languages}.
\newblock In \emph{Findings of the Association for Computational Linguistics:
  ACL-IJCNLP 2021}, pages 4693--4703, Online. Association for Computational
  Linguistics.

\bibitem[{Hermann et~al.(2015)Hermann, Kocisky, Grefenstette, Espeholt, Kay,
  Suleyman, and Blunsom}]{NIPS2015_afdec700}
Karl~Moritz Hermann, Tomas Kocisky, Edward Grefenstette, Lasse Espeholt, Will
  Kay, Mustafa Suleyman, and Phil Blunsom. 2015.
\newblock \href
  {https://proceedings.neurips.cc/paper_files/paper/2015/file/afdec7005cc9f14302cd0474fd0f3c96-Paper.pdf}
  {Teaching machines to read and comprehend}.
\newblock In \emph{Advances in Neural Information Processing Systems},
  volume~28. Curran Associates, Inc.

\bibitem[{Houlsby et~al.(2019)Houlsby, Giurgiu, Jastrzebski, Morrone,
  De~Laroussilhe, Gesmundo, Attariyan, and Gelly}]{houlsby}
Neil Houlsby, Andrei Giurgiu, Stanislaw Jastrzebski, Bruna Morrone, Quentin
  De~Laroussilhe, Andrea Gesmundo, Mona Attariyan, and Sylvain Gelly. 2019.
\newblock \href {https://proceedings.mlr.press/v97/houlsby19a.html}
  {Parameter-efficient transfer learning for {NLP}}.
\newblock In \emph{Proceedings of the International Conference on Machine
  Learning}, Proceedings of Machine Learning Research, pages 2790--2799.

\bibitem[{Hu et~al.(2022)Hu, Shen, Wallis, Allen-Zhu, Li, Wang, Wang, and
  Chen}]{hu2022lora}
Edward~J Hu, Yelong Shen, Phillip Wallis, Zeyuan Allen-Zhu, Yuanzhi Li, Shean
  Wang, Lu~Wang, and Weizhu Chen. 2022.
\newblock \href {https://openreview.net/forum?id=nZeVKeeFYf9} {Lo{RA}: Low-rank
  adaptation of large language models}.
\newblock In \emph{International Conference on Learning Representations}.

\bibitem[{Huang et~al.(2023)Huang, Liu, Lin, Pang, Du, and
  Lin}]{huang2023lorahub}
Chengsong Huang, Qian Liu, Bill~Yuchen Lin, Tianyu Pang, Chao Du, and Min Lin.
  2023.
\newblock \href {http://arxiv.org/abs/2307.13269} {Lorahub: Efficient
  cross-task generalization via dynamic lora composition}.

\bibitem[{Ilharco et~al.(2023)Ilharco, Ribeiro, Wortsman, Schmidt, Hajishirzi,
  and Farhadi}]{ilharco2023editing}
Gabriel Ilharco, Marco~Tulio Ribeiro, Mitchell Wortsman, Ludwig Schmidt,
  Hannaneh Hajishirzi, and Ali Farhadi. 2023.
\newblock \href {https://openreview.net/forum?id=6t0Kwf8-jrj} {Editing models
  with task arithmetic}.
\newblock In \emph{The Eleventh International Conference on Learning
  Representations}.

\bibitem[{Izmailov et~al.(2018)Izmailov, Podoprikhin, Garipov, Vetrov, and
  Wilson}]{izmailov2018averaging}
Pavel Izmailov, Dmitrii Podoprikhin, Timur Garipov, Dmitry Vetrov, and
  Andrew~Gordon Wilson. 2018.
\newblock \href {http://arxiv.org/abs/1803.05407} {Averaging weights leads to
  wider optima and better generalization}.
\newblock Conference on Uncertainty in Artificial Intelligence (UAI), 2018.

\bibitem[{Jiang et~al.(2024)Jiang, Sablayrolles, Roux, Mensch, Savary, Bamford,
  Chaplot, de~las Casas, Hanna, Bressand, Lengyel, Bour, Lample, Lavaud,
  Saulnier, Lachaux, Stock, Subramanian, Yang, Antoniak, Scao, Gervet, Lavril,
  Wang, Lacroix, and Sayed}]{jiang2024mixtralexperts}
Albert~Q. Jiang, Alexandre Sablayrolles, Antoine Roux, Arthur Mensch, Blanche
  Savary, Chris Bamford, Devendra~Singh Chaplot, Diego de~las Casas, Emma~Bou
  Hanna, Florian Bressand, Gianna Lengyel, Guillaume Bour, Guillaume Lample,
  Lélio~Renard Lavaud, Lucile Saulnier, Marie-Anne Lachaux, Pierre Stock,
  Sandeep Subramanian, Sophia Yang, Szymon Antoniak, Teven~Le Scao, Théophile
  Gervet, Thibaut Lavril, Thomas Wang, Timothée Lacroix, and William~El Sayed.
  2024.
\newblock \href {http://arxiv.org/abs/2401.04088} {Mixtral of experts}.

\bibitem[{Jin et~al.(2023)Jin, Ren, Preotiuc-Pietro, and
  Cheng}]{jin2023dataless}
Xisen Jin, Xiang Ren, Daniel Preotiuc-Pietro, and Pengxiang Cheng. 2023.
\newblock \href {https://openreview.net/forum?id=FCnohuR6AnM} {Dataless
  knowledge fusion by merging weights of language models}.
\newblock In \emph{The Eleventh International Conference on Learning
  Representations}.

\bibitem[{Karimi~Mahabadi et~al.(2021)Karimi~Mahabadi, Ruder, Dehghani, and
  Henderson}]{karimi-mahabadi-etal-2021-parameter}
Rabeeh Karimi~Mahabadi, Sebastian Ruder, Mostafa Dehghani, and James Henderson.
  2021.
\newblock \href {https://doi.org/10.18653/v1/2021.acl-long.47}
  {Parameter-efficient multi-task fine-tuning for transformers via shared
  hypernetworks}.
\newblock In \emph{Proceedings of the 59th Annual Meeting of the Association
  for Computational Linguistics and the 11th International Joint Conference on
  Natural Language Processing (Volume 1: Long Papers)}, pages 565--576, Online.
  Association for Computational Linguistics.

\bibitem[{Lester et~al.(2021)Lester, Al-Rfou, and
  Constant}]{lester-etal-2021-power}
Brian Lester, Rami Al-Rfou, and Noah Constant. 2021.
\newblock \href {https://doi.org/10.18653/v1/2021.emnlp-main.243} {The power of
  scale for parameter-efficient prompt tuning}.
\newblock In \emph{Proceedings of the 2021 Conference on Empirical Methods in
  Natural Language Processing}, pages 3045--3059, Online and Punta Cana,
  Dominican Republic. Association for Computational Linguistics.

\bibitem[{Li et~al.(2022{\natexlab{a}})Li, Gururangan, Dettmers, Lewis,
  Althoff, Smith, and Zettlemoyer}]{li2022branchtrainmerge}
Margaret Li, Suchin Gururangan, Tim Dettmers, Mike Lewis, Tim Althoff, Noah~A.
  Smith, and Luke Zettlemoyer. 2022{\natexlab{a}}.
\newblock \href {http://arxiv.org/abs/2208.03306} {Branch-train-merge:
  Embarrassingly parallel training of expert language models}.

\bibitem[{Li et~al.(2022{\natexlab{b}})Li, Gururangan, Dettmers, Lewis,
  Althoff, Smith, and Zettlemoyer}]{btm-suchin}
Margaret Li, Suchin Gururangan, Tim Dettmers, Mike Lewis, Tim Althoff, Noah~A.
  Smith, and Luke Zettlemoyer. 2022{\natexlab{b}}.
\newblock \href {https://doi.org/10.48550/ARXIV.2208.03306}
  {Branch-train-merge: Embarrassingly parallel training of expert language
  models}.

\bibitem[{Li and Liang(2021)}]{li-liang-2021-prefix}
Xiang~Lisa Li and Percy Liang. 2021.
\newblock \href {https://doi.org/10.18653/v1/2021.acl-long.353} {Prefix-tuning:
  Optimizing continuous prompts for generation}.
\newblock In \emph{Proceedings of the 59th Annual Meeting of the Association
  for Computational Linguistics and the 11th International Joint Conference on
  Natural Language Processing (Volume 1: Long Papers)}, pages 4582--4597,
  Online. Association for Computational Linguistics.

\bibitem[{Lin(2004)}]{lin-2004-rouge}
Chin-Yew Lin. 2004.
\newblock \href {https://aclanthology.org/W04-1013} {{ROUGE}: A package for
  automatic evaluation of summaries}.
\newblock In \emph{Text Summarization Branches Out}, pages 74--81, Barcelona,
  Spain. Association for Computational Linguistics.

\bibitem[{Littell et~al.(2017)Littell, Mortensen, Lin, Kairis, Turner, and
  Levin}]{littell-etal-2017-uriel}
Patrick Littell, David~R. Mortensen, Ke~Lin, Katherine Kairis, Carlisle Turner,
  and Lori Levin. 2017.
\newblock \href {https://aclanthology.org/E17-2002} {{URIEL} and lang2vec:
  Representing languages as typological, geographical, and phylogenetic
  vectors}.
\newblock In \emph{Proceedings of the 15th Conference of the {E}uropean Chapter
  of the Association for Computational Linguistics: Volume 2, Short Papers},
  pages 8--14, Valencia, Spain. Association for Computational Linguistics.

\bibitem[{Matena and Raffel(2021)}]{fisheravg}
Michael Matena and Colin Raffel. 2021.
\newblock \href {https://doi.org/10.48550/ARXIV.2111.09832} {Merging models
  with fisher-weighted averaging}.

\bibitem[{Narayan et~al.(2018)Narayan, Cohen, and
  Lapata}]{narayan-etal-2018-dont}
Shashi Narayan, Shay~B. Cohen, and Mirella Lapata. 2018.
\newblock \href {https://doi.org/10.18653/v1/D18-1206} {Don{'}t give me the
  details, just the summary! topic-aware convolutional neural networks for
  extreme summarization}.
\newblock In \emph{Proceedings of the 2018 Conference on Empirical Methods in
  Natural Language Processing}, pages 1797--1807, Brussels, Belgium.
  Association for Computational Linguistics.

\bibitem[{Nenkova and McKeown(2011)}]{INR-015}
Ani Nenkova and Kathleen McKeown. 2011.
\newblock \href {https://doi.org/10.1561/1500000015} {Automatic summarization}.
\newblock \emph{Foundations and Trends in Information Retrieval}, pages
  103--233.

\bibitem[{Ortiz-Jimenez et~al.(2023)Ortiz-Jimenez, Favero, and
  Frossard}]{ortizjimenez2023task}
Guillermo Ortiz-Jimenez, Alessandro Favero, and Pascal Frossard. 2023.
\newblock \href {http://arxiv.org/abs/2305.12827} {Task arithmetic in the
  tangent space: Improved editing of pre-trained models}.

\bibitem[{Pfeiffer et~al.(2021)Pfeiffer, Kamath, R{\"u}ckl{\'e}, Cho, and
  Gurevych}]{pfeiffer-etal-2021-adapterfusion}
Jonas Pfeiffer, Aishwarya Kamath, Andreas R{\"u}ckl{\'e}, Kyunghyun Cho, and
  Iryna Gurevych. 2021.
\newblock \href {https://doi.org/10.18653/v1/2021.eacl-main.39}
  {{A}dapter{F}usion: Non-destructive task composition for transfer learning}.
\newblock In \emph{Proceedings of the 16th Conference of the European Chapter
  of the Association for Computational Linguistics: Main Volume}, pages
  487--503, Online. Association for Computational Linguistics.

\bibitem[{Pfeiffer et~al.(2023)Pfeiffer, Ruder, Vulic, and
  Ponti}]{pfeiffer2023modulardeeplearning}
Jonas Pfeiffer, Sebastian Ruder, Ivan Vulic, and Edoardo~Maria Ponti. 2023.
\newblock \href {https://doi.org/10.48550/ARXIV.2302.11529} {Modular deep
  learning}.
\newblock \emph{arXiv preprint}.

\bibitem[{Pfeiffer et~al.(2020)Pfeiffer, Vuli{\'c}, Gurevych, and
  Ruder}]{pfeiffer-etal-2020-mad}
Jonas Pfeiffer, Ivan Vuli{\'c}, Iryna Gurevych, and Sebastian Ruder. 2020.
\newblock \href {https://doi.org/10.18653/v1/2020.emnlp-main.617} {{MAD-X}:
  {A}n {A}dapter-{B}ased {F}ramework for {M}ulti-{T}ask {C}ross-{L}ingual
  {T}ransfer}.
\newblock In \emph{Proceedings of the 2020 Conference on Empirical Methods in
  Natural Language Processing (EMNLP)}, pages 7654--7673, Online. Association
  for Computational Linguistics.

\bibitem[{Ponti et~al.(2023)Ponti, Sordoni, Bengio, and
  Reddy}]{ponti-etal-2023-combining}
Edoardo~Maria Ponti, Alessandro Sordoni, Yoshua Bengio, and Siva Reddy. 2023.
\newblock \href {https://doi.org/10.18653/v1/2023.eacl-main.49} {Combining
  parameter-efficient modules for task-level generalisation}.
\newblock In \emph{Proceedings of the 17th Conference of the European Chapter
  of the Association for Computational Linguistics}, pages 687--702, Dubrovnik,
  Croatia. Association for Computational Linguistics.

\bibitem[{Radford et~al.(2019)Radford, Wu, Child, Luan, Amodei, and
  Sutskever}]{radford2019language}
Alec Radford, Jeff Wu, Rewon Child, David Luan, Dario Amodei, and Ilya
  Sutskever. 2019.
\newblock \href
  {https://d4mucfpksywv.cloudfront.net/better-language-models/language-models.pdf}
  {Language models are unsupervised multitask learners}.
\newblock \emph{OpenAI Blog}.

\bibitem[{Raffel et~al.(2020)Raffel, Shazeer, Roberts, Lee, Narang, Matena,
  Zhou, Li, and Liu}]{c4}
Colin Raffel, Noam Shazeer, Adam Roberts, Katherine Lee, Sharan Narang, Michael
  Matena, Yanqi Zhou, Wei Li, and Peter~J. Liu. 2020.
\newblock \href {http://jmlr.org/papers/v21/20-074.html} {Exploring the limits
  of transfer learning with a unified text-to-text transformer}.
\newblock \emph{Journal of Machine Learning Research}.

\bibitem[{Rebuffi et~al.(2017)Rebuffi, Bilen, and Vedaldi}]{rebuffi}
Sylvestre-Alvise Rebuffi, Hakan Bilen, and Andrea Vedaldi. 2017.
\newblock \href
  {https://proceedings.neurips.cc/paper/2017/file/e7b24b112a44fdd9ee93bdf998c6ca0e-Paper.pdf}
  {Learning multiple visual domains with residual adapters}.
\newblock In \emph{Advances in Neural Information Processing Systems}.

\bibitem[{Scialom et~al.(2020)Scialom, Dray, Lamprier, Piwowarski, and
  Staiano}]{scialom-etal-2020-mlsum}
Thomas Scialom, Paul-Alexis Dray, Sylvain Lamprier, Benjamin Piwowarski, and
  Jacopo Staiano. 2020.
\newblock \href {https://doi.org/10.18653/v1/2020.emnlp-main.647} {{MLSUM}: The
  multilingual summarization corpus}.
\newblock In \emph{Proceedings of the 2020 Conference on Empirical Methods in
  Natural Language Processing (EMNLP)}, pages 8051--8067, Online. Association
  for Computational Linguistics.

\bibitem[{Touvron et~al.(2023)Touvron, Lavril, Izacard, Martinet, Lachaux,
  Lacroix, Rozière, Goyal, Hambro, Azhar, Rodriguez, Joulin, Grave, and
  Lample}]{touvron2023llama}
Hugo Touvron, Thibaut Lavril, Gautier Izacard, Xavier Martinet, Marie-Anne
  Lachaux, Timothée Lacroix, Baptiste Rozière, Naman Goyal, Eric Hambro,
  Faisal Azhar, Aurelien Rodriguez, Armand Joulin, Edouard Grave, and Guillaume
  Lample. 2023.
\newblock \href {http://arxiv.org/abs/2302.13971} {Llama: Open and efficient
  foundation language models}.

\bibitem[{{\"U}st{\"u}n et~al.(2020){\"U}st{\"u}n, Bisazza, Bouma, and van
  Noord}]{ustun-etal-2020-udapter}
Ahmet {\"U}st{\"u}n, Arianna Bisazza, Gosse Bouma, and Gertjan van Noord. 2020.
\newblock \href {https://doi.org/10.18653/v1/2020.emnlp-main.180} {{UD}apter:
  Language adaptation for truly {U}niversal {D}ependency parsing}.
\newblock In \emph{Proceedings of the 2020 Conference on Empirical Methods in
  Natural Language Processing (EMNLP)}, pages 2302--2315, Online. Association
  for Computational Linguistics.

\bibitem[{Vidoni et~al.(2020)Vidoni, Vulić, and
  Glavaš}]{vidoni2020orthogonal}
Marko Vidoni, Ivan Vulić, and Goran Glavaš. 2020.
\newblock \href {http://arxiv.org/abs/2012.06460} {Orthogonal language and task
  adapters in zero-shot cross-lingual transfer}.

\bibitem[{Vu et~al.(2022)Vu, Barua, Lester, Cer, Iyyer, and
  Constant}]{vu-etal-2022-overcoming}
Tu~Vu, Aditya Barua, Brian Lester, Daniel Cer, Mohit Iyyer, and Noah Constant.
  2022.
\newblock \href {https://aclanthology.org/2022.emnlp-main.630} {Overcoming
  catastrophic forgetting in zero-shot cross-lingual generation}.
\newblock In \emph{Proceedings of the 2022 Conference on Empirical Methods in
  Natural Language Processing}, pages 9279--9300, Abu Dhabi, United Arab
  Emirates. Association for Computational Linguistics.

\bibitem[{Wang et~al.(2018)Wang, Singh, Michael, Hill, Levy, and
  Bowman}]{wang-etal-2018-glue}
Alex Wang, Amanpreet Singh, Julian Michael, Felix Hill, Omer Levy, and Samuel
  Bowman. 2018.
\newblock \href {https://doi.org/10.18653/v1/W18-5446} {{GLUE}: A multi-task
  benchmark and analysis platform for natural language understanding}.
\newblock In \emph{Proceedings of the 2018 {EMNLP} Workshop {B}lackbox{NLP}:
  Analyzing and Interpreting Neural Networks for {NLP}}, pages 353--355,
  Brussels, Belgium. Association for Computational Linguistics.

\bibitem[{Wortsman et~al.(2022)Wortsman, Ilharco, Gadre, Roelofs,
  Gontijo-Lopes, Morcos, Namkoong, Farhadi, Carmon, Kornblith, and
  Schmidt}]{pmlr-v162-wortsman22a}
Mitchell Wortsman, Gabriel Ilharco, Samir~Ya Gadre, Rebecca Roelofs, Raphael
  Gontijo-Lopes, Ari~S Morcos, Hongseok Namkoong, Ali Farhadi, Yair Carmon,
  Simon Kornblith, and Ludwig Schmidt. 2022.
\newblock \href {https://proceedings.mlr.press/v162/wortsman22a.html} {Model
  soups: averaging weights of multiple fine-tuned models improves accuracy
  without increasing inference time}.
\newblock In \emph{Proceedings of the 39th International Conference on Machine
  Learning}.

\bibitem[{Xue et~al.(2021)Xue, Constant, Roberts, Kale, Al-Rfou, Siddhant,
  Barua, and Raffel}]{xue-etal-2021-mt5}
Linting Xue, Noah Constant, Adam Roberts, Mihir Kale, Rami Al-Rfou, Aditya
  Siddhant, Aditya Barua, and Colin Raffel. 2021.
\newblock \href {https://doi.org/10.18653/v1/2021.naacl-main.41} {m{T}5: A
  massively multilingual pre-trained text-to-text transformer}.
\newblock In \emph{Proceedings of the 2021 Conference of the North American
  Chapter of the Association for Computational Linguistics: Human Language
  Technologies}, pages 483--498, Online. Association for Computational
  Linguistics.

\bibitem[{Yadav et~al.(2023)Yadav, Tam, Choshen, Raffel, and
  Bansal}]{yadav2023tiesmerging}
Prateek Yadav, Derek Tam, Leshem Choshen, Colin Raffel, and Mohit Bansal. 2023.
\newblock \href {http://arxiv.org/abs/2306.01708} {Ties-merging: Resolving
  interference when merging models}.
\newblock In \emph{Advances in Neural Information Processing Systems}.

\bibitem[{Yunis et~al.(2022)Yunis, Patel, Savarese, Vardi, Frankle, Walter,
  Livescu, and Maire}]{yunis2022on}
David Yunis, Kumar~Kshitij Patel, Pedro Henrique~Pamplona Savarese, Gal Vardi,
  Jonathan Frankle, Matthew Walter, Karen Livescu, and Michael Maire. 2022.
\newblock \href {https://openreview.net/forum?id=TZQ3PKL3fPr} {On convexity and
  linear mode connectivity in neural networks}.
\newblock In \emph{OPT 2022: Optimization for Machine Learning (NeurIPS 2022
  Workshop)}.

\bibitem[{Zhang et~al.(2023{\natexlab{a}})Zhang, Chen, Liu, and
  He}]{zhang2023composing}
Jinghan Zhang, Shiqi Chen, Junteng Liu, and Junxian He. 2023{\natexlab{a}}.
\newblock \href {http://arxiv.org/abs/2306.14870} {Composing
  parameter-efficient modules with arithmetic operations}.
\newblock In \emph{Advances in Neural Information Processing Systems}.

\bibitem[{Zhang et~al.(2023{\natexlab{b}})Zhang, Han, Liu, Gao, Zhou, Hu, Yan,
  Lu, Li, and Qiao}]{zhang2023llamaadapterefficientfinetuninglanguage}
Renrui Zhang, Jiaming Han, Chris Liu, Peng Gao, Aojun Zhou, Xiangfei Hu, Shilin
  Yan, Pan Lu, Hongsheng Li, and Yu~Qiao. 2023{\natexlab{b}}.
\newblock \href {http://arxiv.org/abs/2303.16199} {Llama-adapter: Efficient
  fine-tuning of language models with zero-init attention}.

\end{thebibliography}
\bibliographystyle{acl_natbib}

\newpage
\appendix

\section{Appendix}
\label{sec:appendix}

\subsection{Are PEFT methods competitive to full fine-tuning of \LLM{}?}

We present the performance of LoRA and Kronecker, two PEFT methods, when used to fine-tune \LLM{} on summarization in 11 languages of XLSum in Table \ref{table:indomain}. We compare their performance to full fine-tuning of \LLM{}. 

Fine-tuning the model with LoRA results in summarization scores that are only $0.4$ ROUGE points below full fine-tuning, while fine-tuning with Kronecker provides a performance similar to full fine-tuning (i.e., just $0.2$ points worse than full fine-tuning). Based on this finding, we conclude that using PEFT methods to fine-tuning \LLM{}, a state-of-the-art LLM, is largely impactful, as in our experiments LoRA for example trains only 0.2\% of the model's parameters
whereas fully tuning the LLM requires updates on 100\% of the model's parameters.

\subsection{XLSum$_\text{seen}$ Dataset}

We are showing the dataset sizes of XLSum$_\text{seen}$ in Table \ref{table:xlsumseen}. 
% This is a section in the appendix.
\begin{table*}[]
\resizebox{\textwidth}{!}{
\begin{tabular}{lrrrrrrrrrrrr}
\toprule
 % & \multicolumn{10}{c}{\textbf{10 Evaluation Domains}}               &         \\
  \textbf{Method}  & Ar & Bn&  En &Id & Ja & Ko & Ru & Sw & Te & Th & Tr &  \multicolumn{1}{c}{Avg} \\  \midrule 
LoRA & 23.4 & 27.6 & 23.5 & 25.0 & 33.6 & 30.4 & 21.3 & 27.1  & 26.9 & 24.7 & 25.3 & 26.2 \\
Multi-LoRA & 23.0 & 27.8 & 22.5 & 24.6 & 34.0 & 30.4 & 20.8 & 27.1 & 27.8 & 25.1 & 24.9 & 26.2 \\ \midrule 
Kronecker & 23.4 & 27.7 & 23.1 & 24.8 & 34.6 & 31.2 & 21.6 & 27.1 & 27.4 & 24.8 & 25.2 & 26.4 \\

Multi-Kronecker & 22.8 & 27.5 & 22.5 & 24.9 & 34.7 & 31.2 & 20.8 & 27.5 & 27.6 & 24.8 &  25.2 & 26.3\\ \midrule
Full fine-tuning & 23.9 & 28.1 & 22.6 & 25.3 & 34.8 & 30.4 & 21.8 & 27.0 & 28.2 & 24.6 & 25.4 & 26.6\\ 
   \bottomrule   
\end{tabular}
} \caption{\textbf{Parameter-efficient fine-tuning vs Full fine-tuning}. Rouge (ROUGE-2 spm) in-domain scores on the $\text{XLSum}_{seen}$ test set. }\label{table:indomain}

\end{table*}

\begin{table}[]
\resizebox{0.9\columnwidth}{!}{
\begin{tabular}{lrr}
\toprule
 Language & Lang code & Dataset size\\ \midrule   Arabic & ar & 38k\\
 Bengali &  bn & 8k \\
English & en &306k \\
Indonesian & id & 38k \\
Japanese & ja &7k \\
Korean & ko & 4k \\
Russian & ru & 62k \\
Swahili & sw & 8k \\
Telugu & te & 10k \\
Thai & th & 7k \\
Turkish & tr & 27k \\ 
\bottomrule   
\end{tabular}

} \caption{Languages in XLSum seen and dataset sizes (training).}\label{table:xlsumseen}
\end{table}

\end{document}